\documentclass[sigconf]{acmart}

\usepackage{booktabs} % For formal tables
\usepackage{latexsym}
\usepackage{url}
\usepackage{subcaption}

\usepackage{multirow}
\hypersetup{draft}
\begin{document}

\copyrightyear{2019}
\acmYear{2019}
\setcopyright{acmlicensed}
\acmConference[ICAIL '19]{Seventeenth International Conference on Artificial
Intelligence and Law}{June 17--21, 2019}{Montreal, QC, Canada}
\acmBooktitle{Seventeenth International Conference on Artificial Intelligence and
Law (ICAIL '19), June 17--21, 2019, Montreal, QC, Canada}
\acmPrice{15.00}
\acmDOI{10.1145/3322640.3326714}
\acmISBN{978-1-4503-6754-7/19/06}
%\title{Recognizing Stance and Arguments from Public Comments in eRulemaking}
%\title{Hierarchical Neural Argument Identification from eRulemaking Public Comments}
\title{Argument Identification in Public Comments from eRulemaking}
%hierarchical
%\titlenote{Produces the permission block, and
%  copyright information}
%\subtitle{Extended Abstract}
%\subtitlenote{The full version of the author's guide is available as
%  \texttt{acmart.pdf} document}

 \author{Vlad Eidelman}
 \orcid{0000-0002-0386-0853}
 \affiliation{%
   \institution{FiscalNote Research}
   \city{Washington}
   \state{DC}
 }
 \email{vlad@fiscalnote.com}

 \author{Brian Grom}
 \affiliation{%
   \institution{FiscalNote Research}
   \city{Washington}
   \state{DC}
 }
 \email{brian@fiscalnote.com}


\begin{abstract}

Administrative agencies in the United States receive millions of comments each year concerning proposed agency actions during the eRulemaking process.
%In recent years, as both the ease of participation and interest in rulemaking have grown, there has been an explosion of public participation,  %
%%As these comments are submitted by a wide range of stakeholders %, including affected companies, advocacy groups, and interested individuals,
These comments represent a diversity of %%perspectives and
arguments in support and opposition of the proposals.
While agencies are required to identify and respond to substantive comments, they have struggled to keep pace with the volume of %%submitted
information.
%%, and it has become infeasible for them to manually review and analyze all received comments.
%While agencies are required to review and respond to these comments, in a painstakingly manual manner, as part of the notice-and-comment process.
%%Computational tools are needed to assist in the reviewing process by automatically identifying pertinent arguments. %% being made for or against the proposed regulation that can be addressed by the agency.

In this work we address the tasks of identifying argumentative text, %%at the sentence level,
classifying the type of argument claims employed, and determining the stance of the comment. % at the document level.
%automatically analyzing comment text to identify whether a span of text contains an argument supporting or opposing the regulation, and classifying/identifying the type of argument.
First, we propose a taxonomy of argument claims based on an analysis of thousands of rules and millions of comments. %employed by commenters . % how to analyze comments, taxonomy of themes (claims)
%propose set of aspect entities comments express opinion on to regulatory agency
Second, we collect and semi-automatically bootstrap annotations to create a dataset of millions of sentences with argument claim type annotation at the sentence level.
Third, we build a system for automatically determining argumentative spans and claim type using our proposed taxonomy in a hierarchical classification model.
\end{abstract}



%\keywords{argument identification, stance detection, erulemaking}

\maketitle
\section{Introduction}

%lots of work on stance
Public comments submitted %%to administrative agencies
during the notice-and-comment portion of eRulemaking have the potential to considerably alter the course of a regulation. In recent years as the public has become more  %%aware and
engaged in the process, the number of submissions has exponentially increased into the millions. %% of comments.
While the Administrative Procedures Act (APA) has been interpreted to require agencies to identify and respond to substantive comments~\cite{fed_rule}, agencies have struggled to keep pace with the volume of submitted information, and it has become infeasible for them to manually review and analyze
all received comments~\cite{Farina}. Therefore, there is a clear need for computational tools that can assist in the drafting and reviewing process by %%analyzing the comments and
automatically identifying pertinent arguments being made for or against the proposed regulation
that can be addressed by the agency, or suggesting argument types to employ to increase the substantivity of a comment.

The task of opinion mining is to automatically process a textual document to detect subjective information, such as an opinion, sentiment or stance~\cite{Pang:2008:OMS:1454711.1454712}. Stance detection is a type of opinion mining where given a piece of text and a target entity, the goal is to automatically identify whether the author of the text holds a supporting, opposing, or neutral position with respect to the target. Stance detection has been applied to a variety of domains, including news~\cite{DBLP:journals/corr/abs-1804-00982} and social media~\cite{Wei:2018:MSD:3209978.3210145}. In the case of comments, the authors' target is the proposed regulation.

A related task, sentiment analysis deals with assigning a positive, negative, or neutral polarity to subjective text~\cite{Pang:2008:OMS:1454711.1454712}.
Both stance detection and sentiment analysis can be carried out at the document level, sentence level, or at the sub-sentential aspect level.
Aspect-based sentiment analysis (ABSA) is a more fine-grained version where different opinion polarity can be expressed with regard to different entities and their attributes, even within the same sentence~\cite{DBLP:journals/corr/abs-1804-00982}.
Argument mining shares subtasks with sentiment analysis and stance detection, and attempts to analyze argumentation structures in text by identifying argumentative language, classifying claims and premises, and identifying argumentative relations~\cite{Lawrence:2017:UAS:3106680.3032989,Stab:2017:PAS:3160785.3160790,Lippi:2016:AMS:2909066.2850417}.

While there have been previous efforts to perform stance detection, sentiment analysis, and argumentation mining on regulatory comments~\cite{livermore,Kwon:2007:ICS:1248460.1248473,Park:2015:TMP:2746090.2746118}, this domain poses a number of unique challenges.

%Typical applications of stance detection (refs) more clear stance, regulatory comments different for several reasons.
First, as regulatory comments are submitted by a wide range of
stakeholders,
%with different levels of sophistication and subject matter knowledge,
including affected companies, advocacy groups, and interested individuals,
they represent a diversity of perspectives and arguments in support and opposition of the proposals.
%they represent a diversity of arguments in support and opposition, and %% of the proposals.
Commenters have different levels of sophistication and subject matter knowledge, resulting in varying comment length, depth, and writing styles, ranging from several sentences from a public submission
to many dense pages from a law firm or trade association.
%Future work: Important to understand the importance the comment may hold for the agency in it's final rulemaking.
%Different authors place different levels of effort into authoring.
%Most tweets/reviews follow a typical format.

Second, a regulation will often have multiple sections and
affect several distinct
extant administrative rules and code sections. Thus, the comment may hold a nuanced position, containing multiple arguments that either support or oppose different pieces of the rule.

Third, comments may not explicitly mention the regulation or its content, yet still express a stance.
For example, authors may question the rulemaking process, the agency, the legislation that created the statutory authority, or agree with another comment.%, along with a host of other legitimate ways of expressing a substantive opinion regarding the regulation.%, despite not needing to mention it or its content.

Finally, regulations will almost inevitably be promulgated. Thus, it is much more useful for the agency to understand \textit {why} an author is for or against a regulation, so they can either acknowledge the points or modify the final rule, than the overall stance.
% but if strong points are made, they should at least acknolwedge, if not make changes to the final regulation.
A %%binary (support/oppose), or
single stance is unable to provide the substantive understanding %%of the public
necessary for an agency to meaningfully consider the comments.%%, nor for an individual to get an understanding of how the rule may evolve.

From the agencies' perspective, comments such as the ones at the top of Table~\ref{tbl:comm_example}, without proper justifying arguments are unlikely to be treated substantively. On the other hand, comments like the ones at the bottom of Table~\ref{tbl:comm_example}, especially if they are from prominent authors, or repeated across many commenters, are more likely to result in an agency response~\cite{livermore}.

 \begin{table*}
   \caption{ Example of substantive and non-substantive comments with different supporting arguments. \label{tbl:comm_example}}

  \begin{center}
  \begin{tabular}{|p{2.75cm}|p{2.5cm}|p{11cm}|} \hline   Substantivity  & Argument Type & Example Comment \\ \hline\hline
 \multirow{2}{*} {Non-substantive}  & Explicit Support & This sounds good. Up front at least. It's been a while since this country has passed any major environmental movements, and I definitely agree with one like this. \\ \hline
  & Likely Support & Thank you, thank you for your courage and foresight. Change must happen immediately to save the planet.\\ \hline
 \hline
 \multirow{3}{*} {Substantive } & Likely Opposition & Although I approve of the proposed expansion on the definition of medical sources, I cannot support nor agree with the elimination of all deference to treating physicians. \\ \hline
  & Likely Opposition   Disputed \newline Information & While we support certain HRSA proposals, such as those related to telemedicine and group purchasing organization (GPO) exceptions, we have serious concerns about others, which often appear to be without basis or justification. \\ \hline
 &Explicit Opposition &  After a thorough review of the CPP, we believe the wisest course of action is for EPA to withdraw the proposed rule and abandon its costly agenda to regulate carbon dioxide under the Clean Air Act. The reasons for our position are straightforward. EPA's proposed rule: \\
 &Overreach & Is illegal, stretching far beyond the narrow boundaries of Section 111(d) of the Clean Air Act;\\
 &Burdensome & Imposes high costs for no meaningful benefits. Electricity will be more expensive for small business owners and entrepreneurs, which will slow economic growth, harm competitiveness, and destroy jobs. There will be no effect on global temperatures and climate change.\\
 &Burdensome & Threatens the reliability of the nation's bulk electric power system, which raises the prospect of blackouts and brownouts, which can in turn increase operating expenses and uncertainty, as well as reduce output and revenues.\\ \hline
 \end{tabular}
  \end{center}
  \end{table*}

In this work, we are interested in the tasks of identifying argumentative text at the sentence level, classifying the type of argument claims employed, and determining the stance of the comment. % at the document level.
%automatically analyzing comment text to identify whether a span of text contains an argument supporting or opposing the regulation, and classifying/identifying the type of argument.
To that end, our contributions are as follows. First, we propose a taxonomy of argument claims based on an analysis of thousands of rules and millions of comments. %employed by commenters . % how to analyze comments, taxonomy of themes (claims)
%propose set of aspect entities comments express opinion on to regulatory agency
Second, we collect and semi-automatically bootstrap annotations to create a dataset of millions of sentences with argument claim type annotation at the sentence level.
Third, we build a system for automatically determining argumentative spans and claim type. We show that while this is a difficult task, using our proposed taxonomy in a hierarchical classification strategy results in the best performance for argument claim identification.

%analysis the different themes

%
%Results can be used to summarize the response to a proposed rule

%some about the process, some about the regulation itself?

%Target regulation may or may not be referred to in the comment. Some comments explicitly target proposals in the rule, others target the agency, or Congressional act, or ...

\section{Related Work}
\label{sec:related}

Computational regulatory comment analysis has received significant attention %%from the NLP community and legal scholars, %% for some time.
%%The problems addressed with computational approaches have covered
including identifying stakeholders~\cite{Arguello:2007:BAI:1248460.1248475}, clustering duplicate comments~\cite{Yang:2006:NSN:1146598.1146663}, topic modeling~\cite{livermore}, creating topic ontologies~\cite{Yang:2008:OGL:1367832.1367875}, verifiability of propositions \cite{W14-2105}, automatically classifying the dialogical type of relations~\cite{Lawrence:2017:UAS:3106680.3032989}, theoretical argumentation models~\cite{Park:2015:TMP:2746090.2746118}, and a rule-dependent issue hierarchy~\cite{DBLP:conf/lrec/CardieFRA08}.
%lots of previous work examining specific regulations, and some examining commenting behavior on specific regulations
%opinion mining has traditionally been applied to news, social media, but there have been application to Congressional behavior as well.
%In relation to stance detection, someone can express exclusively positive sentiment while being opposed to the target, or vice versa.

The closest work in spirit to ours may be that of~\citet{Kwon:2006:MTA:1146598.1146649} and~\citet{Kwon:2007:ICS:1248460.1248473} who
perform topic classification, claim identification and 3-way sentence-level stance detection which they refer to as claim classification,
on the claims for one EPA rule. %~\cite{Kwon:2006:MTA:1146598.1146649,Kwon:2007:ICS:1248460.1248473}. %%In relation to our work,
We perform a similar claim identification on a much larger scale, and our claim classification goes beyond stance into specific and generic regulatory claim types.

~\citet{W14-2105} created a dataset of under 10,000 sentences from 1,000 comments submitted on two rules to predict the verifiability of argument propositions.
They remove non-argumentative propositions, which account for only 7\% of their data, and train an SVM classifier using n-gram and various count features.
In a similar direction, ~\citet{Park:2015:TMP:2746090.2746118} propose a theoretical argumentation model for assisting authors in drafting evaluable comments.~\citet{DBLP:conf/lrec/CardieFRA08} create a rule-dependent issue hierarchy and annotate a set of 267 comments from an FTA rule at the sentence level.

~\citet{Lawrence:2017:UAS:3106680.3032989} annotate a corpus of comments from one rule with an argumentative and dialogical structure. They then build models to automatically classify the dialogical type of relations between propositions.

Outside of regulatory comment analysis, opinion and argument mining have been applied to a variety of text, including movie reviews, product reviews, student essays, and legal writing~\cite{Pang:2008:OMS:1454711.1454712,D16-1171,Stab:2017:PAS:3160785.3160790,conf/icail/PalauM09a}.

We assume the author holds a single supporting or opposing stance at the document level. %%of either opposing or supporting the target proposed regulation. %, and we treat that as the argument claim.
The author may never explicitly state their stance, thus it may be unobserved and need to be inferred from the text. The author may substantiate their stance through the use of various arguments employed at the sentence level, which we consider the claims, i.e. reasons justifying the stance.
So in the argumentation framework, we are interested in claim identification - identifying the argumentative components, and claim classification - classifying the types of argument claims. We leave the final step of identifying premises supporting the claims and their validity or persuasiveness to future work.\footnote{Our task could be framed as a multi-aspect sentiment analysis problem, where we first determine what the pertinent aspects across proposed rules are, and use those to identify the authors sentiment toward specific aspects for a given rule and comment. For instance, if {\bf Burdensome} and {\bf Flexibility} were two aspects, a commenter could believe a new rule creates no additional burden or even that a burden may be removed, and that there is enough flexibility (all positive), or that the rule creates a tremendous economic burden and lacks flexibility (negative). However, given our annotation results that the much more common intent of commenters is to argue with an aspect of the proposed rule, as many more comments complain about the flexibility, clarity, cost, etc. than praise those aspects, we chose to model the problem as one of argument identification and assume either a negative or positive sentiment per aspect type. We believe this is a reasonable assumption and leave the task of automatically learning the sentiment with respect to each of the proposed aspects to future work.}

\section{Data}

% types of arguments commenters use to ,

% automatically assign them at the sentence level, and classify the comment stance at the document level.

%%In order to perform argument identification and classification,
We constructed a new dataset of comment documents annotated with argument claim types at the sentence level. In this section we explain the process we used to create our dataset and semi-automatically bootstrap annotations through weak supervision.

\subsection{Collection}

First, we selected dockets with more than 50 and fewer than 1,000 comments submitted to regulations.gov\footnote{https://www.regulations.gov/ is cross-agency initiative hosted by the EPA as part of the eRulemaking Program providing public access to federal regulatory content.} from the first date available through submission on January 1, 2019. %%For those dockets %%using the publicly available API
We collected all available comment data, including associated metadata and text.\footnote{On regulations.gov authors are able to submit comment text directly through a text field or as an attachment. As attachments come in various different formats, and many are not machine readable and thus require additional OCR capabilities, in order to reduce the noisiness of the data we only include direct submissions in the corpus. We leave the inclusion of attachment to future work.}

As previous work has indicated~\cite{Yang:2006:NSN:1146598.1146663}, there are a substantial number of form comments submitted with large portions of repeated content. In order to have a set of unique sentences, we remove duplicates by performing a fuzzy similarity match above a threshold of 95\% similar at the sentence level.\footnote{This also prevents us from artificially inflating performance by having duplicate content in the training and test sets.}
The resulting dataset is composed of approximately 1.8 million sentences from 2095 separate dockets.
Table~\ref{tbl:agency_dist} shows the distribution of comments over the top represented agencies.

 \begin{table}
  \begin{center}
    \caption{ Agencies with over 30 rules in the corpus. \label{tbl:agency_dist}}

  \begin{tabular}{|l|r|} \hline Agency  & Number of rules    \\ \hline

 EPA & 360 \\
 CMS & 155\\
 FDA & 141\\
 OSHA & 112\\
 FWS & 104\\
 FAA & 98\\
 NOAA & 77\\
 APHIS & 70\\
 USCG & 65\\
 DOT & 52\\
 ED & 50\\
 FMCSA &  49 \\
 NHTSA & 44 \\
 CFPB&   34 \\
 HUD  & 34 \\
 FHWA &  32 \\
 PHMSA & 31 \\ \hline

  \end{tabular}
  \end{center}
  \end{table}

%Selected dockets with more than 50 and fewer than 1,000 comments,

\subsection{Claim Taxonomy}

To create the taxonomy of argument claims, three regulatory subject matter experts manually reviewed approximately 20,000 comments to identify the types of claims that are directly used to argue for or against a proposed rule. %%and identified distinct %%repeated argument %%claim types. While there may be many kinds of opinion or subjective statements made in a comment, the objective of our work is to identify the broad types of claims that are directly used to argue for or against a proposed rule.
This resulted in the definition of 12 specific and 4 generic argument claims, presented below:%in Table~\ref{tbl:arg_type_examples} along with example sentences from the corpus. %% identified by our annotation model (described below) as belonging to each argument type.

%%below. The specific claims are:

{\bf Burdensome:} Too arduous or burdensome, either financially or administratively, or requesting to mitigate a burden.

  {\bf Not Sufficient Time:} Too short a time frame for either commenting or actual implementation.

  {\bf Lacks Flexibility:} Overly excessive, harsh, or prescriptive.

  {\bf Conflicting Interests:} Favorable to one party at the expense of others, creating a conflict of interest. %%(or the appearance of it,)
%  or being in conflict with another agency's opinion or authority.

  {\bf Disputed Information:} Containing assertions, studies or premises that are flawed or have no factual basis. %%in the commenter's opinion.

  {\bf Legal Challenge:} Contrary to prior judicial rulings, an invitation to judicial review, or likely to increase the chance of litigation.

  {\bf Overreach:} Outside the agency's scope of authority, or outside the legislative intent of the authorizing statute.

  {\bf Requests Clarification:} Points that need clarification, are unclear or outright demand modification.

  {\bf Seeks Exclusion:} Individuals or organizations receiving special treatment or exemptions from the general application of the rule.

  {\bf Lacks Clarity:} Needing more definition, description, or general clarification.

  {\bf Too Broad:} Too sweeping in nature. The rule looks to have unintended impact. %%on individuals and organizations,
%  or contains vague language that may encourage open interpretation.

  {\bf Too Narrow:} Too restrictive or specific in nature.  %%If worded too specifically, the rule may appear irrelevant, without being applied to the very individuals or organizations it was designed to affect.

% \noindent In addition to the above, the four generic types for support and opposition are:

  {\bf Explicit Support (Opposition):} Supporting (opposing) the rule or a portion explicitly. Usually the comment directly references the rule or regulation in a supporting (opposing) statement.
% % "I am writing to urge you to make the proposed rule final."
% %"If this proposal is not adopted, every new device will be treated as capped rental, placing negative pressure on innovation."

 {\bf Likely Support (Opposition):} Supporting (opposing) the rule or a portion implicitly.

\noindent Tables~\ref{tbl:arg_type_examples} and ~\ref{tbl:arg_type_examples2} present several example sentences from the corpus from each argument type. %% identified by our annotation model (described below) as belonging to each argument type.
 We further group the argument claims into a hierarchy, presented in Table~\ref{tbl:arg_data}.\footnote{All argument types are assumed to justify, and thus fall under, one stance, and some argument types further qualify another argument.}
%"MPSC is not opposed to the FCC setting end goals and timelines for all types of intercarrier compensation, including intrastate access."
%"I strongly urge the EPA to move forward with your proposed framework to reduce carbon pollution from existing power plants."

%% {\bf Explicit Opposition} Opposing the rule or a portion explicitly. Usually the comment directly references the rule or regulation in an opposition statement.
% "Please do not go forward with the proposed rule."
%"I strongly urge the agency to withdraw this thoroughly flawed rule..."
%"I also urge you to not to finalize these regulatory changes..."

%%{\bf Likely Opposition} Arguments involving the commenter opposing the rule or a portion implicitly.
%"The FCC is supposed to be working for and protecting the public, not doing the work of corporations."
%"The agency should not restrict these sustainable methods of farming without data showing an actual, verified increased rate of foodborne illness."

%%Tables~\ref{tbl:arg_type_examples} and~\ref{tbl:arg_type_examples2} present several example sentences
%%from the corpus identified by our annotation model (described below) as belonging to each argument type.
%%We further group the argument claims into a hierarchy, presented in Table~\ref{tbl:arg_data}. All argument types are assumed to justify, and thus fall under, one stance, and some argument types further qualify another argument.

\begin{table*}
 \caption{ Example sentences identified by the rule-based model as belonging to the respective argument type. \label{tbl:arg_type_examples}}

 \begin{center}
 \begin{tabular}{|p{3cm}|p{13.7cm}|} \hline   Argument Type & Example Sentences \\ \hline

%\multirow{4}{*}
\multirow{4}{*}{\bf Burdensome}& The current rule, which requires separate filings for each legal entity, unnecessarily increases the administrative burden on CMRS contributors and USAC.\\
& These multiple levels of ineligibility cause an additional burden for CRNAs to have access to this technology in order to report quality measures electronically.\\
&This proposal ignores the major adverse economic impacts it would create, as well as the economic costs that would negate the net benefit of any effort to expand service at current total support levels.\\
& We are concerned, however, that the IFR will be so costly that it challenges many of our members survival due to the increased costs, paperwork burdens and administrative hurdles now imposed by the rule.\\ \hline

\multirow{4}{*}{\bf Lacks Flexibility} &  One deficiency to trigger re competition is overly harsh. \\
&       180 days of service requirement is too prescriptive. \\
&       Should the physician know about the referral ... yes but there needs to be flexibility in how to proceed.\\
&       Many growers do not have the ability to spread dry granular products or the cost is prohibitively high. \\ \hline

\multirow{4}{*}{\bf Not Sufficient Time } & Most importantly, we would request additional time to collaborate with APHIS to identify further efficiencies that might be realized in lieu of a rate increase.\\
%& Sprint points out that the timeframe of 60-days or less is not sufficient to include many towers deployed on a temporary basis.\\
& Six months is not an adequate timeframe in which to begin managing episodes that will be subject to downside risk.\\
%&Given that chronic Hyalella azteca tests are conducted over a duration of 42 days, the IEPA concurs that a retest prior to the March 30, 2016 deadline is infeasible. \\ \hline

\multirow{4}{*} {\bf Conflicting Interests} & This gives them an incentive to abuse their ISP relationship to give an advantage to their content over other content providers they do not own.\\
& A prime contractor could also gain a competitive advantage over a subcontractor in future business dealings by having obtained such sensitive information.\\
& Dealing with state licensing rules there are times when the state rule is in conflict with the Federal rule. \\
%& There is a conflict of interest for the major fiber network operators to sell transport at reasonable prices to ISPs that appear to compete with the retail operations of these big network operators, even if the intent is to serve customers in rural areas that the big networks choose not to serve.\\
& The proposed standard conflicts with the soil fertility and crop nutrient management practice standard of the NOP regulations (7 C.F.R. 205.203.) \\\hline
\multirow{3}{*}{\bf Disputed Information} & There is no evidence to support the hypothesis that allowing cross ownership will increase with the quantity of diversity of news available in smaller markets, as hypothesized by the Chief Economist.\\
&The mythological notion that regulations are bad for jobs and the economy has been repeatedly debunked, but it keeps coming back.\\
&Specifically, such suggestions are inconsistent with Section 254(e)'s mandate that all universal service support be explicit. \\\hline
\multirow{4}{*} {\bf Legal Challenge } & The current proposal is the product of a flawed process and will likely result in the same types of policy conflicts and legal battles which have plagued wolf management in the past.\\
& The new 201.3(a) will inevitably invite costly litigation between growers and processors over settled issues of case law, resulting in injury to both parties.\\
& ...adopting such provisions would lead to endless litigation between incumbents and geographic licensees.\\
& This proposal is fraught with issues for the nonprofit sector and creates real legal concerns. \\ \hline

\multirow{4}{*}{\bf Overreach}  & The proposed rule ignores Congressional intent and Supreme Court rulings, and impermissibly expands Federal jurisdiction.\\
& Efforts to prohibit these services represent an unnecessary Overreach \\
& This dual-track of enforcement violates the letter and spirit of the humane slaughter law and exacerbates problems repeatedly raised in government oversight reports.\\
&The proposed regulations would considerably expand federal authority into what is currently a state-level and institution-level jurisdiction.\\\hline
\end{tabular}
 \end{center}
 \end{table*}

\begin{table*}
 \caption{ Example sentences identified by the rule-based model as belonging to the respective argument type. \label{tbl:arg_type_examples2}}

 \begin{center}
 \begin{tabular}{|p{3cm}|p{13.7cm}|} \hline   Argument Type & Example   \\ \hline

\multirow{4}{*} {\bf Requests Clarification } & As drafted, this language is ambiguous and the body of the final rule should clearly express the EPA's intention regarding which entity or state can claim the benefit of renewable energy dispatch.\\
& The proposed rule should provide examples of unwelcome conduct concerning the use of religious symbols...\\
&This proposed rule needs to be clarified to apply only to new construction.\\
&The satellite industry has been proceeding with concrete plans to launch networks using the 37 and 39 GHz bands, but satellite investment cannot continue under the regulatory uncertainty that would result from the proposed hybrid auction and license approach.\\ \hline

\multirow{4}{*}{\bf Seeks Exclusion } &  You must remove this clause and give religious relief organizations the freedom to truly help the children who need their help. \\
& Can wording be updated to allow school districts exemption from the 5 year requirement since we secure more recent updates? \\
& Soaps are also listed under 40 CFR180.950 that grants tolerance exemptions for both minimal risk active and inert ingredients. \\
& We are asking for an exemption for treated foundation seed stock to be used by licensed seed breeders when there is no untreated seed stock commercially available to them. \\ \hline

\multirow{3}{*}{\bf Lacks Clarity} &    There should be a clear instruction on how to measure for school readiness. \\
&   However we find the wording in this section of the rule confusing and potentially misleading. \\
&   USDA should clearly identify when a covered commodity is the same. \\ \hline

 \multirow{4}{*}{\bf Too Broad }  & I believe the proposed rule is overly broad and vague as written and as a result will actually harm a lot of horses for which this practice is not used.\\
 & Throughout the rule, broad terms are frequently used, leaving many requirements open to interpretation.\\
 & For example, too broad a definition could effectively result in most healthcare facilities being declared a medical device manufacturer.\\
& The Commission also should narrow its proposal at Appendix C paragraph 270 concerning the ability of the calling party service provider to choose whether to interconnect directly or indirectly with the called party.\\ \hline
\multirow{4}{*}{\bf Too Narrow} & The current rule is too narrow and outdated, and threatens the retirement security of millions of Americans.\\
&The Commission also should consider expanding its definition of failed and failing stations.\\
&This appears to be an overly restrictive criterion.\\
&The alternative is to cast the exemption too narrowly, as the Commission has done, in an effort to avoid inclusion of additional entities. \\ \hline

 \multirow{2}{*}{\bf Explicit Support }  &  I am writing to urge you to make the proposed rule final. \\
& If this proposal is not adopted, every new device will be treated as capped rental, placing negative pressure on innovation. \\ \hline

\multirow{3}{*}{\bf Likely Support} & We appreciate USDA's willingness to help state agencies maximize the audit funds. \\
& I support the change that will require each grantee to meet key quality indicators to receive renewal.\\
& I strongly urge the EPA to move forward with your proposed framework to reduce carbon pollution from existing power plants. \\ \hline

\multirow{3}{*} {\bf Explicit Opposition} & Please do not go forward with the proposed rule.\\
&I strongly urge the agency to withdraw this thoroughly flawed rule.\\
&I also urge you to not to finalize these regulatory changes. \\ \hline

\multirow{4}{*}{\bf Likely Opposition} & We only ask that the panel consider the negative effects of some of the statements in the proposal that would hinder quality services.\\
& Rural programs do not benefit from this rule at all. \\
& The agency should not restrict these sustainable methods of farming without data showing an actual, verified increased rate of foodborne illness.\\ \hline

 \end{tabular}
 \end{center}
 \end{table*}

\subsection{Annotation}

%%The goal of annotation was to create a corpus that could be used to train machine learning models to automatically perform claim identification and classification.
In order to scalably annotate a corpus with millions of sentences, %%we need to trade-off complexity and cost with annotation accuracy, thus
we chose a weakly supervised
%human-in-the-loop
semi-automated approach~\cite{alex2017snorkel}.

We start with a set of seed words and phrases identified by the subject matter experts %%per claim type,
and automatically label the data using rule-based labeling functions. The rule-based model is based on a set of rules applied using the CKY~\cite{Cocke:1969:PLC:1097042} context-free parsing algorithm.

We employ a number of discourse, subjectivity, and other lexical cues~\cite{Kwon:2007:ICS:1248460.1248473,Lawrence:2017:UAS:3106680.3032989}, in the labeling criteria of the rules, including semantic orientation, polarity bearing, and semantically similar word lists, such as WordNet, and direct policy mentions. %The final rule-based system is composed of X rules. Table~\ref{tbl:rule_breakdown} shows the number of rules per claim type, and Table~\ref{tbl:rule_examples} shows example rules.

During development of the grammar, after applying the current version of the labeling functions we cluster sentences with similar subjective phrases %%using known and computed word and phrase similarities,
and %%use the new words and phrases to manually
expand the seed set and above-mentioned cues per claim type. We then relabel the relevant clusters with the updated model and iterate.% the analysis until sentences either have a label or are not believed to contain further claims.

%Two annotators were trained to evaluate sentence annotations according to claim taxonomy. They found an average of 80\% of labels across claim types were correct. %The inter-annotator reliability is fairly high.

%lack clarity 303/1050 = 28
%conflict interest
%seek exclusion = 151/459

Table~\ref{tbl:arg_data} shows the stance and argument type hierarchy alongside the distribution of argument types in the final annotated corpus. As previous work has also shown~\cite{DBLP:conf/lrec/CardieFRA08}, we see that the corpus is highly imbalanced toward non-argumentative sentences, as neutral sentences make up over 87\%. This is expected as many parts of comments are expressing non-argumentative statements, such as greetings, pleasantries, or other opinions that are not directly used to argue for or against a proposed rule.% providing evidence to support an argument.

Of the opinion-bearing sentences, {\bf likely support} and {\bf likely opposition} are the most common across all argument types, %We have a much richer taxonomy for opposing arguments than supporting,
%There are many more opposition than supporting ones,
with opposition making up over 75\%. %%, and the {\bf likely opposition} type making up almost half of that.
This is by design, as one of the primary motivations for this work is to enable an understanding of the specific types of disagreement present in the comments.%% in order for the agency to take proper action.
%%It is useful to know the extent to which comments agree in order to validate the rules usefulness, but there is little for the agency to take action on.
%\squeezeupp
%\vspace{-1mm}
\footnote{To relate this to a quote from Anna Karenina by Leo Tolstoy: "\textit{Happy families are all alike; every unhappy family is unhappy in its own way.}" When commenters agree with the proposal, they generally express a general sense of satisfaction, while those who oppose have many reasons why.}

Beyond our current task, this corpus can be utilized in a number of ways, such as describing the differences in arguments received by different agencies, correlating the rule content to the types of arguments received, and predicting if and what arguments receive attention from the agency or result in modifications to the final rule. We leave these analyses to future work.

\begin{table}
 \caption{ Argument hierarchy and number of occurences in the corpus. \label{tbl:arg_data}}
% \squeezeupp
 \begin{center}
 \begin{tabular}{|p{1.25cm}|p{1.5cm}|l|r|} \hline Stance  & { Argument    } &  {Argument }  & Number \\ \hline
Neutral & &&1613085 \\ \hline\hline
 \multirow{15}{*} {Opposition} %& & & \\\cline{2-4}
& Explicit & &19012\\ \cline{2-4}
& Likely  && 89396\\
 &&Burdensome  & 12001  \\
 &&---Lacks Flexibility &1754\\
 &&---Not Sufficient Time &3820\\
 &&Conflicting Interests&2050\\
&&Disputed Information&20943\\
&&Legal Challenge &1785 \\
 &&Overreach &2982\\
 &&Requests Clarification &5021\\
 &&---Lacks Clarity &9848\\
 &&---Seeks Exclusion &4119\\
 &&Too Broad & 478\\
 &&Too Narrow &2187\\ \hline\hline

\multirow{2}{*}{Support} % && & \\ \cline{2-4}
 & Explicit & &14647\\ \cline{2-4}
 &Likely  &&42701\\ \hline
 \end{tabular}
 \end{center}
% \squeezeupp
% \squeezeupp

 \end{table}

 \section{Methods}
 \label{sec:methods}

This section describes our comparison of approaches to building a system for automatic claim identification and classification. %Since there may be multiple claims in a document, we can approach the problem in several different ways and decompose it into several binary and multiclass classification problems.

First, we employ a flat classification approach and jointly perform claim identification and classification as a multiclass problem where we classify all claim types directly and ignore the hierarchy.

Second, we separate the tasks, and first tackle the claim identification as a binary problem of identifying non-argumentative (neutral) versus argumentative sentences. Then, given a sentence has been identified as argumentative, we pose a multiclass problem of classifying the arguments into different claim types.

Finally, we split the multiclass problem above into another binary classification problem, where we identify the stance of the sentence, as opposing or supporting, and then perform a separate multiclass classification on each of the sets of supporting and opposing argument types separately. The latter results in a top-down hierarchical classification approach~\cite{Silla:2011:SHC:1937796.1937884}, where we have separate models for claim identification, stance detection, and two separate models for the supporting and opposing argument type classification.

%imbalanced class problem, most statements are neutral

%neutral vs all (multiclass)
%neutral vs argument (binary)
%-->argument all (multiclass)
%-->support vs opposition (binary)
%----->support all (multiclass)
%----->opposition all (multiclass)

\subsection{Model Description}

We perform the binary and multiclass classification described above with a standard bag of n-grams based linear model in the form of a logistic regression model. % and Support Vector Machine (SVM)~\cite{}.
While neural models are increasingly popular, linear models are simple, fast to train, and have been shown to have strong performance on text classification and sentiment analysis tasks~\cite{conf/acl/WangM12,ZhangW15b}.
We use the \texttt{scikit-learn}~\cite{sklearn} implementation. The feature space is binary valued n-gram presence; whether a unigram or bigram occurs in the sentence. We retain the most frequent 30k n-grams. % across the training and dev sets.
We further augment each sentence with a weighted average of its component word vectors followed by PCA~\cite{arora2016simple} using pre-trained Google word2vec 300-dimensional word vectors~\cite{mikolov2013distributed}.
%, and implemented NBSVM based on the interpolated version in~\newcite{conf/acl/WangM12}.
In addition, we train a fastText~\cite{joulin2016bag} model using 50-dimensional word vectors with unigrams and bigrams. %, which is based on an distributed word representations.
fastText represents a document as the average of its word vectors and trains a linear model on that representation. It %is fast and
has been shown to have perform competitively with more complex deep neural models~\cite{DBLP:journals/corr/LeCD17}. %We train the fastText models using 50-dimensional word vectors with unigrams and bigrams.

%We additionally use a CNN model architecture similar to that proposed in ~\cite{ZhangW15b,DBLP:journals/corr/Kim14f}.
%The model consists of an embedding layer, followed by 3 convolutional layers with kernel window sizes of 3,4,5 and 100 filters, a global max-pooling layer on each filter whose output is concatenated and fed into a fully connected layer, and finally a classification layer with a softmax activation function.
%Although deeper and more complex neural models have been proposed for text classification, ~\cite{DBLP:journals/corr/LeCD17} showed that the so-called shallow-and-wide CNN is consistently a top performer.

%The the neural models, w use pre-trained Google word2vec 300-dimensional word vectors~\cite{}. Words outside the pre-trained vocabulary are randomly initialized.
%Each sentence is padded to a maximum length $l$, and represented as a concatenation of its word embeddings.
%Represent each
%\vspace{-1mm}
\section {Experiments}

For evaluation we utilize 70\% of the data for training, with the remainder serving as a test and dev set, retaining the relative proportions of each argument type in each set.
%We carry out experiments to validate the performance of the alternative modeling frameworks described above.
Due to the imbalanced nature of the data, accuracy and micro-averaged F$_1$ can be optimized by predicting only the majority class (e.g. neutral for claim identification), so we optimize and measure results on macro-averaged F$_1$. %%, which averages the performance on each class individually. %%, as opposed to micro- which aggregates all classes to calculate the performance.
We use Bayesian hyperparameter optimization~\cite{NIPS2011_4443} implemented in \texttt{hyperopt}~\cite{bergstra_hyperopt-proc-scipy-2013} %for our sequential model-based optimization~\cite{bergstra_hyperopt-proc-scipy-2013}
to select the best hyperparameters for each model on the dev set. We set class weights during training to be inversely proportional to the occurrence of the class in the training data.  %We used the tree-structured Parzen Estimator (TPE) algorithm implemented in \texttt{hyperopt} for our sequential model-based optimization~\cite{bergstra_hyperopt-proc-scipy-2013}.
%\vspace{-1mm}
 \subsection{Results}
 \label{sec:results}

%neutral vs support/oppose arguments
Results for claim identification are presented in Table~\ref{tbl:res_bin_neut_arg}. %We evaluated the model in two settings.
We first assess the ability to learn to identify arguments in a balanced setting, with the number of instances of the neutral class downsampled. %The performance
%%in classifying sentences
Both models achieve a macro-F$_1$ score of around 0.90. %% to equal that of the argumentative class.
%%we are interested in the identification performance
In the imbalanced setting where neutral sentences are highly prevalent, %%, we test the performance of the model on all the data.
%% In the imbalanced setting,
 the neutral performance remains high
%%at an F$_1$ of 0.92,
but argument identification drops to 0.62-0.65 F$_1$, %%mainly driven by the significant drop in recall, where only half the argument sentences are labeled as such,
resulting in 0.77-0.80 macro-F$_1$.
%in balanced case with neutral adding up to other categories

 %a balanced setting with the number of instances of the neutral class downsampled to equal that of the argumentative class, and the real imbalanced %setting represented in the data.
%How well can we distinguish neutral from argument sentences

%how well support from oppose

Given sentences that are argumentative, we first perform binary stance detection, whose results are presented in Table~\ref{tbl:res_bin_supp_opp}. This task is also imbalanced, with more opposition examples than support. This likely explains why we perform better on opposition, with 0.89-0.90 F$_1$.%%, and lower on support, with a significantly lower recall.

Alterntively %Given sentences that are known to be argumentative,
Table~\ref{tbl:res_mult_all_noneut} (claim-neutral) shows results for performing multiclass claim classification on all - oppose and support - argument types together. %% in a multiclass setting.
The overall performance is 0.60-0.64 F$_1$, with {\bf likely opposition}, {\bf legal challenge}, and {\bf seeks exclusion} performing the best, while {\bf lacks flexibility}, {\bf too narrow}, and {\bf too broad} are worst.

Given a known stance for the sentence, Table~\ref{tbl:res_mult_all_noneut} (supp v. opp) shows results for performing multiclass claim classification for oppose and support argument types separately. Somewhat surprisingly we see stronger overall performance on support than oppose, with a macro-F$_1$ of 0.85 for support and 0.65-0.70 for oppose. On an argument level,  {\bf likely support} and  {\bf likely opposition} both do well, while  {\bf too broad},  {\bf too narrow}, and  {\bf lacks flexibility} perform the worst.%%, similar to multiclass setting with the exception of {\bf likely support}.

%sThe latter would result in a top-down hierarchical classification approach~\cite{Silla:2011:SHC:1937796.1937884}, where we have separate models for claim identification, stance detection, and two separate models for the supporting and opposing argument type classification.
%Alternatively, we could employ a flat classification approach and jointly perform claim identification and classification as a multiclass problem where we classify all claim types directly and ignore the hierarchy.
In all of the above except claim identification, we assume we know which sentences are argumentative, or what their stance is. However, in a real-world setting we do not. Table~\ref{tbl:res_mult_all_noneut} (claim+neutral) presents performance for %modeling the problem as a multiclass with
both argumentative and neutral sentences. %% from the start.
As expected, we achieve a significantly lower macro-F$_1$ as the arguments' classification performance drops, while the neutral class remains in the 0.90s. Least affected are {\bf conflicting interests}, {\bf burdensome}, and {\bf requests clarification}.

Finally, taking advantage of the proposed hierarchical structure of the argument taxonomy, we train the claim identification, stance detection, and separate stance-based claim classification models as before, where we have knowledge of the true argumentative sentences and stance, and augment the claim+neutral data with each of their output probabilities. We use this augmented data to build an ensemble claim classification model that in addition to the original feature space has the predicted probabilities of each of the above models as input features. Performance for the ensemble model is shown in Table~\ref{tbl:res_mult_all_noneut} (claim+ensemble). This model achieves a macro-F$_1$ improvement of 4\% over claim+neutral to 0.5. %The highest individual increases in performance are for {\bf seeks exclusion}, whose precision jumped from 0.47 to 0.71, and {\bf explicit opposition} and {\bf explicit support}, both of which increased at least 0.11 points.

\begin{table}
\caption{Claim identification and stance detection.}
\begin{subtable}[h]{.99\columnwidth}
 \begin{center}
 \begin{tabular}{|c|c|c|c|c||c|c|c|} \hline
%\cline{3-8}
& &\multicolumn{3}{|c|}{LR+w2v} & \multicolumn{3}{|c|}{fastText} \\ \hline
 Setting   & Type & { R    } &    P  & F$_1$ & { R    } &    P  & F$_1$ \\ \hline

\multirow{3}{*} {bal} &    Arg   &  0.90  &    0.88   &   0.89  & 0.91& 0.90 & 0.90\\
       &    Neut &       0.89   &   0.90   &   0.89  & 0.90& 0.91& 0.90\\\cline{2-8}
      &        Macro-Ave &0.89  &     0.89    &  0.89 & 0.90& 0.90& 0.90\\ \hline
\multirow{3}{*} {imbal}  & Arg &  0.51     &  0.79   &   0.62  & 0.71& 0.59& 0.65  \\
      &        Neut & 0.97  &     0.89    &  0.92& 0.94 & 0.97 & 0.95\\ \cline{2-8}
&        Macro-Ave & 0.74  &     0.84    &  0.77& 0.83 & 0.78 & 0.80\\

      \hline
 \end{tabular}
 \end{center}
 \caption{ Claim identification: neutral vs. argumentative classification for balanced and imbalanced settings. \label{tbl:res_bin_neut_arg}}
 \end{subtable}

\begin{subtable}[h]{.95\columnwidth}
 \begin{center}
 \begin{tabular}{|c|c|c|c||c|c|c|} \hline
& \multicolumn{3}{|c|}{LR+w2v} & \multicolumn{3}{|c|}{fastText} \\ \hline
  Stance &  { R   } &    P & F$_1$ &  { R   } &    P & F$_1$\\ \hline

 Opposition     &  0.92 &     0.85  &    0.88  & 0.89 & 0.92 & 0.90\\
         Support     &  0.61  &   0.76  &    0.68   & 0.71 & 0.63 & 0.67 \\\hline
        Macro-Ave &   0.76    &  0.80 &    0.78 & 0.80 & 0.77 & 0.78\\ \hline
 \end{tabular}
 \end{center}
 \caption{  Stance detection for opposition vs. support. \label{tbl:res_bin_supp_opp}}
 \end{subtable}
% \squeezeup
% \squeezeup
 \end{table}

% \begin{table}
%         \centering
%         \caption{Multiprogram sets}
%         \label{multiprogram}
%         \begin{tabular}{c|c|c|c|c|c|c|c|c|}
%             \cline{2-9}
%              & \multicolumn{8}{|c|}{Sets}\\
%             \cline{2-9}
%              & 1 & 2 & 3 & 4 & 5 & 6 & 7 & 8\\
%             \hline
%             \multicolumn{1}{|c|}{astar} & & * &  & * &  &  & * &\\
%             \hline
%         \end{tabular}
%     \end{table}

\begin{table*}
\centering %\small%\footnotesize
\caption{ Claim classification of argument types.% in  settings.
\label{tbl:res_mult_all_noneut}}
%\squeezeupp
\setlength{\tabcolsep}{2.8pt} % Default value: 6pt
\begin{tabular}{|c|c|c|c||c|c|c||c|c|c||c|c|c||c|c|c||c|c|c||c|c|c|}\hline
& \multicolumn{12}{|c||}{LR+w2v} & \multicolumn{9}{|c|}{fastText} \\ \hline
%\cline{2-24}
& \multicolumn{3}{|l|}{claim-neutral} & \multicolumn{3}{|l|}{supp v. opp} & \multicolumn{3}{|l|}{claim+neutral} & \multicolumn{3}{|l||}{claim+ensemble} & \multicolumn{3}{|l|}{claim-neutral} & \multicolumn{3}{|l|}{supp v. opp} & \multicolumn{3}{|l|}{claim+neutral}  \\
\hline
Type  & { R    } &    P  & F$_1$ & { R    } &    P  & F$_1$ & { R    } &    P  & F$_1$ & { R    } &    P  & F$_1$  &{ R    } &    P  & F$_1$ & { R    } &    P  & F$_1$ & { R    } &    P  & F$_1$ \\ \hline
           Burdensome   &    0.61   &   0.77   &   0.68   &  0.69   &0.81   & 0.74  & 0.43& 0.66 &0.52 & 0.48   &   0.71   &   0.57  &0.65 & 0.67& 0.66 & 0.73 & 0.71 & 0.72 & 0.52 & 0.44 & 0.47\\
Conflict Int.   &    0.63   &   0.75   &   0.68   & 0.70& 0.80&0.75  &0.50 &0.65 &0.56 & 0.52    &  0.69   &   0.59 & 0.74 & 0.62 & 0.68 & 0.77 & 0.67 & 0.72 & 0.62 &      0.44   &   0.52 \\
 Disputed Info   &    0.65   &   0.72   &   0.68   &0.70 &0.75 &0.72 & 0.29& 0.52&0.37 & 0.33   &   0.52  &    0.40 & 0.66 & 0.65 & 0.66 & 0.70 & 0.67 & 0.69 & 0.44 &     0.31     & 0.36\\
  Explicit Opp   &    0.59   &   0.66   &   0.62   & 0.64&0.70 & 0.67 & 0.37&0.60 &0.46 & 0.43  &    0.64  &    0.51& 0.66 & 0.56 &0.61 & 0.71 &0.62 & 0.66 & 0.52 &      0.42     & 0.47  \\
     Explicit Supp   &    0.62   &   0.73   &   0.67   & 0.73&0.83 &0.77  & 0.44&0.66 &0.53 & 0.48   &   0.72  &    0.58 & 0.71 & 0.66 & 0.69 & 0.81 & 0.75 & 0.78 & 0.64  &    0.51  &    0.56\\
        Lacks Clarity   &    0.58   &   0.68   &   0.62  & 0.63& 0.70& 0.66 & 0.25& 0.48& 0.33 & 0.32 &     0.50  &    0.39 & 0.65 & 0.56 & 0.60 &0.68   &   0.60   &   0.64  & 0.42   &   0.28   &   0.34 \\
    Lacks Flexibility   &   0.45    &  0.53    &  0.49   & 0.54& 0.61& 0.58& 0.22& 0.36& 0.27 & 0.30  &    0.46   &   0.36 & 0.53 & 0.38 & 0.44 & 0.63    &  0.43  &   0.51 &  0.32  &    0.17   &   0.22\\
    Likely Opp   &    0.78   &   0.68   &   0.73  & 0.87& 0.78& 0.82& 0.42& 0.48& 0.45 &  0.40    &  0.52  &    0.45& 0.69 & 0.76 & 0.72 & 0.78 &     0.86   &   0.82 &  0.48      &0.40   &   0.43\\
       Likely Support   &    0.62   &   0.52   &   0.57   & 0.94& 0.90& 0.92&0.28 &0.41 &0.33 & 0.29  &    0.44 &     0.35 & 0.57 & 0.57 & 0.57 & 0.92 & 0.94 & 0.93 &  0.38    &  0.29   &   0.33 \\
    Legal Challenge     &  0.69     & 0.78    &  0.73     &0.67 &0.78 &0.72 & 0.40 &0.61 &0.48  &  0.42   &   0.70  &    0.53 &0.72 & 0.56 & 0.63 & 0.70   &   0.57  &    0.63 & 0.46   &   0.32     & 0.38 \\
      Not Suff. Time &      0.60  &    0.75  &    0.67   & 0.67& 0.73& 0.70 & 0.33& 0.56 & 0.41& 0.38    &  0.61  &    0.47  & 0.66 & 0.60 & 0.62  &0.74     & 0.63 &     0.68 & 0.51  &   0.32   &   0.40\\
            Overreach     &  0.59   &   0.75    &  0.66     & 0.63&0.71 &0.67 & 0.35&0.55 &0.43 & 0.41    & 0.59   &   0.48 & 0.66 & 0.59 & 0.62&0.71   &   0.58  &    0.64  & 0.53   &   0.36   &   0.43 \\
Req. Clarification    &   0.58  &    0.70   &   0.63    & 0.72&0.79 &0.76 & 0.42& 0.61& 0.49& 0.45    &  0.66   &   0.54 & 0.65 & 0.55 & 0.60 & 0.78    &  0.66   &   0.72 & 0.55    &  0.42   &   0.47 \\
      Seeks Exclusion     &  0.68   &   0.86    &  0.76      & 0.78& 0.90&0.84 &0.47 &0.75 &0.58& 0.46     & 0.79 &     0.58& 0.75 & 0.71 & 0.73 & 0.82  &   0.77   &   0.79 & 0.57   &   0.42  &    0.48\\
            Too Broad     &  0.50   &   0.60    &  0.54      & 0.49&0.60 &0.54&0.25 &0.41 &0.32 & 0.37  &    0.50  &    0.42 & 0.59 & 0.32 & 0.41 & 0.48   &   0.27 &     0.35 & 0.26   &   0.14  &    0.19 \\
           Too Narrow     &  0.42   &   0.55    &  0.48    & 0.54& 0.61& 0.57& 0.25& 0.41&0.31 & 0.35   &   0.54  &    0.43 & 0.43 & 0.34 & 0.38 & 0.59     & 0.46   &   0.52 & 0.32 &     0.22    &  0.26\\\hline
           Neutral &-&-&-&-&-&-& 0.97& 0.91& 0.94 &  0.96    &  0.90   &   0.93 &-&-&-&-&-&-& 0.94  &    0.97   &   0.96  \\ \hline\hline
            Macro-Ave     &  0.60   &   0.69    &  0.64    &0.75 &0.80 &0.77 & 0.39& 0.57& 0.46 & 0.43   &   0.62    &  0.50 & 0.65 & 0.57 & 0.60 & 0.79 & 0.73 &0.75 &  0.50   &   0.38  &    0.43\\ \hline
%73, 61, 66
%87, 84, 85
%80, 72.5, 75.5

% macro avg       0.66      0.73      0.70
% macro avg       0.83      0.86      0.85     13807

%ft
%macro avg       0.70      0.61      0.65
% macro avg       0.87      0.84      0.85     13807
\end{tabular}
% \squeezeupp

 \end{table*}

 \subsection{Analysis}
 \label{sec:analysis}

To validate whether the semi-automated corpus creation was able to successfully identify the argumentative sentences, and to further understand the ability of the models to learn the argument types and stance, we examine the top weighted n-grams from three models.

First, the top-weighted features from the claim identification model at the top of Table~\ref{tbl:arg_type_top_ngrams}
%we found that the %%top-weighted
%features
for the neutral class have little to do with argumentation, while the features for the argumentative class capture many argumentative bigrams, such as \textit{economic benefit}, \textit{unintended consequences}, and \textit{adversely affect}. %, validating %, and are all related to an argument type.
%This shows
%that our corpus creation strategy was %at least
%able to capture argumentative sentences. % at a high level.
%%Furthermore, we capture many argumentative bigrams, such as \textit{economic benefit}, \textit{unintended consequences}, and \textit{adversely affect}.
Second, %we examine the top-weighted features for supporting and opposing claims for the stance detection model
looking at stance in Table~\ref{tbl:arg_type_top_ngrams}, most of the supporting n-grams express positivity: \textit{no adverse}, \textit{hopeful}, and \textit{not delay}, while the opposing n-grams explicitly disapprove: \textit{not support}, \textit{not approve}, and \textit{please reject}. It is interesting to note that in both cases our model learns negated n-grams, and correctly learns that \textit{no adverse} is supporting, while \textit{not support} is opposing.

Finally, %%we examine the top-weighted features for the claim classification model across all argument types excluding neutral sentences
%in Table~\ref{tbl:arg_type_top_ngrams}
for claim classification we identify n-grams highly indicative of the argument type, such as \textit{science} and \textit{estimates} for {\bf disputed information}, \textit{lawsuit} and \textit{litigation} for {\bf legal challenge}, \textit{artificially} and \textit{unreasonable} for {\bf lacks flexibility}, and \textit{postpone} and \textit{deadline} for {\bf not sufficient time}. This further shows that our corpus creation method was  able to capture meaningful distinction between argumentative and non-argumentative sentences, and identify sentences associated with the argument claim types.% in our proposed taxonomy.

\begin{table*}%[bth]
\caption{ Most highly weighted n-grams by the LR+w2v claim identification, stance detection and claim classification models. %across all argument types without neutral sentences.
\label{tbl:arg_type_top_ngrams}}
%%\begin{table*}
%\squeezeup
%\squeezeupp

 \begin{center}
 \begin{tabular}{|p{2.9cm}|p{13.5cm}|} \hline   Argument Type & Top n-grams   \\ \hline\hline
         Neutral & to regulations, from another, all too, pressure to, as defined, do an, system with, analyzing, and american%, incomes, that beneficiaries, how do, for ensuring, purchase the, you please, so please, do your, also please, please clarify, but please
        \\\hline
        Argumentative & overreach, greatly benefit, unintended consequences, economic benefit, time consuming, %economic benefits%,
        please reconsider%, positive impact, hardships, please change, burdens, extremely high, extremely limited, forcing, lawsuits, adversely affect, more time, forget that, adverse impact, operating costs
        \\ \hline\hline
          Support & not cut, not delay, welcome, economic benefit, commend, appreciates, strengthening, no adverse, helps%, supports, strengthen, positive,  do cut, not place, hopeful, am for, is for
        \\\hline
        Oppose& not support, not approve, not pass, not agree, forcing, lawsuits, not adopt, please add, exceptions%, potential risk, economic impacts, against this, confusing, please change, reject, please remove, reconsider, withdraw, more time
        \\\hline\hline
         Burdensome     &    prohibitive, time consuming, excessive, adverse impact, unnecessary, extremely, hardships, burdens    \\ \hline
Conflicting Interests    &   anti competitive, treatment, greatly benefit, conflict of, unfair, advantage, competitive, conflict, disadvantage     \\\hline
 Disputed Information    &  estimates, study, facts, unintended consequences, theory, studies, claim, evidence, proof, science      \\\hline
  Explicit Opposition    &   not pass, concerned, not address, oppose, urge, object, not agree, cms to, you to, not support     \\\hline
     Explicit Support    &    weaken, please pass, endorse, approved, not restrict, not delay, agree with, supports, not cut, support      \\\hline
        Lacks Clarity    &   forgotten, ambiguous, unsure, explain, uncertainty, specify, clarify, define, confusing, confusion    \\\hline
    Lacks Flexibility    &   artificially, room, inflated, prescriptive, dictate, inflexible, high, flexible, unreasonable, flexibility    \\\hline
    Likely Opposition    &   strict, violate, should consider, disappointed, consider, once again, please reconsider, forcing, potential risk    \\\hline
       Likely Support    &   helps, enacted, strong, help, applaud, positive, appreciates, economic benefit, commend, economic benefits     \\\hline
    Legal Challenge      &   court, suits, to court, sue, sued, legal, suit, lawsuit, litigation, lawsuits    \\\hline
      Not Sufficient Time  &    delay, postpone, timely, more time, schedules, timeframe, timeline, deadline, deadlines, schedule      \\\hline
            Overreach      &  over reach, too far, overstep, right, authority, permission, power, overstepping, overreaching, overreach     \\\hline
Requests Clarification     &   change, please, please change, revision, revise, revising, replace, please add, amend, please remove      \\\hline
      Seeks Exclusion      &  allowing, exclusions, suspension, suspensions, exclusion, exemptions, exceptions, exemption, relief     \\\hline
            Too Broad      &    open to, overly, open, be limited, overly broad, tight, to interpretation, interpretation, specific, broad     \\\hline
           Too Narrow      &  firm, stricter, be expanded, limited, narrow, long overdue, strengthening, strengthen, prohibitive, restrictive      \\\hline
 \end{tabular}
 \end{center}
% \squeezeup
% \squeezeupp
 \end{table*}

\section{Conclusions}

In this work we addressed the related tasks of identifying argumentative text at the sentence level, classifying the type of argument claims employed, and determining the stance of the comment. % at the document level.
%automatically analyzing comment text to identify whether a span of text contains an argument supporting or opposing the regulation, and classifying/identifying the type of argument.

We proposed a taxonomy of argument claims based on %subject matter experts
an analysis of millions of comments that is broadly applicable across different rule types. %employed by commenters . % how to analyze comments, taxonomy of themes (claims)
%propose set of aspect entities comments express opinion on to regulatory agency
We collected and semi-automatically bootstrapped annotations to create a dataset of millions of sentences with argument claim type annotation at the sentence level, and showed that the corpus creation process was successfully able to capture argumentative sentence and our specific proposed claim types. Finally, we presented several ways for automatically determining argumentative spans and claim types, showing the relative performance of each across the different claims. Our results show that while argument classification is a difficult task, our proposed taxonomy can be used to automatically identify and classify claims, with a hierarchical classification strategy achieving the best performance.

%\appendix
%Appendix A

\bibliographystyle{ACM-Reference-Format}
\bibliography{reg_build.bib}

\end{document}